# Qualitative Decision Theory with Sugeno Integrals

Didier Dubois, Henri Prade, Régis Sabbadin
IRIT - Université Paul Sabatier - 31062 Toulouse Cedex (France)
e-mail: {dubois, prade, sabbadin} @irit.fr

## Abstract

This paper presents an axiomatic framework for qualitative decision under uncertainty in a finite setting. The corresponding utility is expressed by a sup-min expression, called Sugeno (or fuzzy) integral. Technically speaking, Sugeno integral is a median, which is indeed a qualitative counterpart to the averaging operation underlying expected utility. The axiomatic justification of Sugeno integral-based utility is expressed in terms of preference between acts as in Savage decision theory. Pessimistic and optimistic qualitative utilities, based on necessity and possibility measures, previously introduced by two of the authors, can be retrieved in this setting by adding appropriate axioms.

## 1 INTRODUCTION

The expected utility criterion was the first to receive axiomatic justifications both in terms of probabilistic lotteries [von Neumann & Morgenstern 1944] for decision under risk and in terms of preference between acts for decision under uncertainty [Savage 1954]. These axiomatic frameworks have been questioned later, challenging some of the postulates leading to the expected utility criterion. Noticeably, [Allais 1953] and later [Ellsberg 1961] laid bare the existence of cases where a systematic violation of the expected utility criterion could be observed. Some of these violations were due to a cautious attitude of decision-makers in front of uncertainty. More recently [Gilboa 1987] and [Schmeidler 1989] have advocated and axiomatized lower and upper expectations expressed by Choquet integrals attached to non-additive measures corresponding to families of probability measures, as a formal approach to utility that accounts for the Ellsberg paradox (see also [Sarin & Wakker 1992]). Some of these generalized expected utility criteria are also generalizations of the Wald criterion for decision under ignorance. The latter suggests that a decision be evaluated by the value of its worst possible consequence. Choquet integrals, especially the lower expectations, are mild versions of the Wald criterion [Wald 1950].

In the framework of Artificial Intelligence, it has been pointed out that information about preference and uncertainty in decision problems cannot always be quantified in a simple way, but only qualitative evaluations can sometimes be attained. As a consequence, the topic of qualitative decision theory is a natural one to consider [Brafman & Tennenholtz 1997]: can we make decision on the basis of qualitative information? In this kind of research the set of states of the world and the set of consequences of actions are often supposed to be finite, contrary to classical frameworks where one of them is infinite. Moreover, giving up the quantification of utility and uncertainty also means that the notion of expectation based on averaging is given up as well. It contrasts with non-expected utilities based on Choquet integrals where the idea of average value of a decision is preserved although it becomes imprecise. Fully giving up quantification of utility and belief leads to a purely symbolic approach where uncertainty is represented by a likelihood relation on events and preference by an ordering on consequences of decisions.

In [Dubois & Prade 1995] two qualitative criteria, an optimistic and a pessimistic one, whose definitions only require a linearly ordered scale for utility and uncertainty are proposed as well as axiomatic justifications for them in the style of von Neumann and Morgenstern and more recently, [Dubois, Prade, & Sabbadin 1997], in the style of Savage. In these qualitative criteria, uncertainty and preferences are captured by possibility distributions ([Zadeh 1978], [Dubois & Prade 1988]), and more precisely in the framework of qualitative possibility theory ([Dubois & Prade 1998], [Dubois 1986]). As pointed out in [Dubois & Prade 1995], the pessimistic (resp. optimistic) criterion is formally equivalent to the degree of necessity (resp. possibility) of a fuzzy event [Dubois & Prade 1988] which can be seen as a particular case of Sugeno integral with respect to a necessity (resp. possibility) measure. This framework is tailored for one-shot (non repeated) decisions, where the decision is evaluated by means of a particular state prevision that belongs to the ones considered as plausible for the decision-maker.

In this paper, enlarging the possibilistic framework of [Dubois, Prade, & Sabbadin 1997], we examine the general case of qualitative evaluations of decision rules, when the uncertainty is modeled by a monotonic set-function, and the criterion is expressed as a Sugeno integral. A Sugeno integral is the qualitative counterpart of a Choquet integral, but it only requires a totally ordered scale to be defined. While classical decision theory uses average to assess the value of decisions under uncertainty, Sugeno integral is a median, which can be viewed as a qualitative counterpart of an expectation. Moreover, the analogy between Choquet integrals and Sugeno integrals is patent due to a similar behavior in the presence of comonotonic acts, where Choquet integrals become additive while Sugeno integral decomposes by the minimum and the maximum. We give an axiomatic decision-theoretic justification for such a criterion, in the case where there is a finite set of states of the world. Before presenting these results, let us recall the axiomatic justification proposed by Savage [Savage 1954] for the expected utility decision criterion.

All the proofs of the theorems of this paper are included in the full version of this paper.

## 2   SAVAGE'S AXIOMATICS FOR EXPECTED UTILITY

Savage has proposed a framework for axiomatizing decision rules under uncertainty where both the uncertainty function and the utility function are derived from first principles bearing on acts. The proposed axioms can be operationally verified by checking how the decision-maker ranks his acts. This section recalls Savage's setting and his axioms for justifying expected utility and probability functions.

In Savage's approach a preference relation $\preceq$ between acts (or decisions) is assumed to be given by a decision-maker. Such a preference relation is observable from the decision-maker's behavior. Acts are defined as functions $\mathbf{f}$ from a state space $S$ to a set $X$ of consequences. Indeed the result of an act depends on the state of the world in which it is performed:

the effect of braking a car depends on the state of the brake. Let us denote $\mathcal{F} = X^S$ the set of potential acts. The set of actually feasible acts is generally only a subset of $\mathcal{F}$. The first assumption made by Savage is that the preference relation on $\mathcal{F}$ is transitive and complete ($\mathbf{g} \preceq \mathbf{f}$ or $\mathbf{f} \preceq \mathbf{g}$):

**Sav 1** *Ranking*: $(\mathcal{F}, \preceq)$ is a complete preorder.

Two particular families of acts are crucial to recover the preference information on consequences and the uncertainty information on the state space $S$: constant acts and binary acts respectively. A *constant act*, denoted $\mathbf{x}$ for $x \in X$ is such that $\forall s \in S, \mathbf{x}(s) = x$. Since $\preceq$ is a complete preorder on $\mathcal{F}$, the set of acts, it is also a complete preorder on the set of constant acts (which can be identified with $X$). Therefore, we can define the following complete preorder $\leq_P$ on $X$:

**Definition 1** *Preference on consequences induced by the ranking of constant acts*: $\forall x, y \in X, \forall s \in S$ if $\mathbf{f}(s) = x$, and $\mathbf{g}(s) = y$, then $x \leq_P y \Leftrightarrow \mathbf{f} \preceq \mathbf{g}$.

In order to avoid the trivial case when there is only one consequence or all consequences are equally preferred, Savage has enforced the following condition:

**Sav 5**[1] *Non triviality*: There exists $x, x' \in X$ such that $x <_P x'$, where $<_P$ is the strict part of the complete preordering on $X$.

The ranking of acts also induces a ranking of events, i.e., subsets of the state space: this is based on the use of binary acts. A *binary act* is an act $\mathbf{f}$ such that there is a set $A \subseteq S$ and two consequences $y <_P x \in X$ where $\mathbf{f}(s) = x$ if $s \in A$, $\mathbf{f}(s) = y$ if $s \in \overline{A}$, where $\overline{A}$ is the complement of $A$. Such a binary act is denoted $xAy$. A partial ordering $\leq_L$ of events can be defined by restricting the complete preordering on acts to binary acts:

**Definition 2** *Relative likelihood of events*: Let $A, B \subseteq S$. Event $A$ is not more likely than event $B$, denoted $A \leq_L B$, if and only if $\forall x, y \in X, y <_P x$, $xAy \preceq xBy$.

Of course relation $\leq_L$ is only a partial preordering. In order to turn it into a complete preordering, Savage proposed the following axiom:

**Sav 4** *Projection from acts over events*: Let $x, y, x', y' \in X, x' <_P x, y' <_P y$. For all $A, B \subseteq S$. $xAx' \preceq xBx' \Leftrightarrow yAy' \preceq yBy'$.

This axiom ensures that for any choice of consequences $x' <_P x$, the restriction of the preordering on acts to binary acts $xAx'$ defines a complete preorder-

---

[1]For the sake of clarity we use Savage's original numbering of axioms

ing of events in a unique way. The notion of binary act is a particular case of a compound act:

**Definition 3** *Compound act:* $\forall A \subseteq S$, $\mathbf{f}A\mathbf{g}$ *is the act defined by:* $\mathbf{f}A\mathbf{g}(s) = \mathbf{f}(s)$ *for all* $s \in A$, *and* $\mathbf{f}A\mathbf{g}(s) = \mathbf{g}(s)$ *for all* $s \in \overline{A}$.

A binary act is thus a compound constant act. Savage has introduced a cancellation property, that makes the following assumption: if two acts give the same results on a subset of states, their relative preference does not depend on what these results are. This is called the sure thing principle and is modeled as follows:

**Sav 2** *Sure thing principle:* Let $\mathbf{f}$, $\mathbf{g}$, $\mathbf{h}$, $\mathbf{h'} \in \mathcal{F}$, let $A \subseteq S$. $\mathbf{f}A\mathbf{h} \preceq \mathbf{g}A\mathbf{h} \Rightarrow \mathbf{f}A\mathbf{h'} \preceq \mathbf{g}A\mathbf{h'}$.

If three acts $\mathbf{f}$, $\mathbf{h}$ and $\mathbf{g}$ are such that $\mathbf{f}A\mathbf{h} \preceq \mathbf{g}A\mathbf{h}$ then $g$ is said to be *conditionally preferred* to act $\mathbf{f}$ on event (a set of states) $A$, denoted $(\mathbf{f} \preceq \mathbf{g})_A$. Clearly, due to the sure thing principle, conditional preference is well-defined, namely the property $(\mathbf{f} \preceq \mathbf{g})_A$ does not depend on the choice of act $\mathbf{h}$. Moreover it is a complete preordering of acts. There is a type of events such that conditioning on them blurs all preferences: null events. An event $A$ is said to be null if and only if $\mathbf{f}A\mathbf{h} \preceq \mathbf{g}A\mathbf{h}$ for any $\mathbf{f}$, $\mathbf{h}$ and $\mathbf{g}$. It can be proved that null events are impossible in the sense that $A \sim_L \emptyset$ if and only if $A$ is null.

The restriction of conditional preference to constant acts must coincide with the preference ordering on consequences (except for null events). This is achieved by the following axiom:

**Sav 3** *Conditioning over constant acts:* Let $x, y \in X$, $A \subseteq S, A$ not null. Let $\mathbf{x}$, $\mathbf{y}$ be the constant acts: $\mathbf{x}(s) = x$ and $\mathbf{y}(s) = y, \forall s \in S$. Then, $(\mathbf{x} \preceq \mathbf{y})_A \Leftrightarrow x \leq_P y$.

Under the above 5 conditions the likelihood relation on events induced by the preference relation on acts is a comparative probability relation, namely it obeys the following characteristic properties: **A1** $\leq_L$ is complete and transitive, **A2** $\emptyset <_L S$ (non-triviality), **A3** $\forall A, \emptyset \leq_L A$ (consistency), **P** if $A \cap (B \cup C) = \emptyset$ then $B \leq_L C$ if and only if $A \cup B \leq_L A \cup C$ (additivity).

The setting proposed by Savage presupposes that the set of states is infinite. This assumption is necessary for the introduction of the following axiom:

**Sav 6**[2] *Quantitative probability:* Let $\mathbf{f}$, $\mathbf{g} \in \mathcal{F}$, such that $\mathbf{f} \prec \mathbf{g}$, let $x \in X$. There exists $\bigcup B_i$ a partition of $S$ such that $\forall i$, $xB_i\mathbf{f} \prec \mathbf{g}$ and $\mathbf{f} \prec xB_i\mathbf{g}$.

Savage proved that a preference relation satisfying

---

[2]This condition is necessary in order to obtain a quantitative representation of the comparative probability ordering.

**Sav 1** to **Sav 6** can be represented by a utility function $u$ from the set of acts to the reals. For any act $\mathbf{f}$, $u(\mathbf{f})$ is the expected utility of the consequences of $\mathbf{f}$ in the sense of a probability distribution on $S$.

## 3  DECISION-MAKING WITH SUGENO INTEGRALS

Consider again the set $\mathcal{F}$ of acts $\mathbf{f}$, mappings from $S$ to $X$. Clearly if we take Savage framework for granted, there is a common evaluation scale for events (i.e. binary acts) and constant acts (just take the set of acts quotiented by the indifference relation). So it is possible to evaluate uncertainty and preference by means of a totally ordered scale $(L, \leq)$ (finite, as we will assume in this paper that $S$ and $X$ are finite). The mapping from the set of consequences to $L$ is a utility function $\mu : X \to L$. It is supposed that the top $1_L$ and the bottom $0_L$ of $L$ are in $\mu(X) = \{\mu(x), x \in X\}$. If not, just add an ideal consequence $x^*$ and a totally bad consequence $x_*$ to $X$. Uncertainty is supposed to be captured by means of a set function $\sigma : S \to L$ which is a monotonic measure (called *fuzzy measure* by [Sugeno 1977]), that is such that: $\sigma(\emptyset) = 0_L$, $\sigma(S) = 1_L$, $A \subseteq B \Rightarrow \sigma(A) \leq \sigma(B)$. This kind of set-function is very general and represents the minimal requirement for the representation of partial belief. Especially the last condition is called *monotonicity*, and is verified by probability measures and most other well-known representations of partial belief (including belief and plausibility functions, necessity and possibility measures...). Then the utility of an act $\mathbf{f}$ can be defined as a Sugeno integral [Sugeno 1977], a qualitative counterpart of expected sum, where the sum is replaced by a *sup* (a *max* in the finite case) and the product by a *min*.

**Definition 4** *Monotonic utility of an act*

$$u_S(\mathbf{f}) = \int_L^{Sug} \mathbf{f} d\sigma = \max_{\lambda \in L} \min(\lambda, \sigma(F_\lambda))$$

*where* $F_\lambda = \{s \in S, \mu(\mathbf{f}(s)) \geq \lambda\}$ *and* $\sigma$ *is a monotonic measure.*

This Sugeno integral is called *monotonic qualitative utility*, and $u_S$ can also be written only by varying the consequence $x$ in $X$, as follows:

**Proposition 1** $u_S(\mathbf{f}) = \max_{x \in X} \min(\mu(x), \sigma(F_x))$ *where* $F_x = \{s \in S, \mu(\mathbf{f}(s)) \geq_P \mu(x)\}$.

Let us analyze the properties of qualitative monotonic utility. First, let us compute the monotonic utilities of some basic acts: for constant acts, it is obvious that $u_S(\mathbf{x}) = \mu(x)$. For compound acts of the form $xAy$, the following property holds:

**Proposition 2** *If $\mu(x) \geq \mu(y)$, then $u_S(xAy) = \max(\mu(y), \min(\mu(x), \sigma(A)))$.*

It follows obviously that: $u_S(x^*Ax_*) = \sigma(A)$. Moreover $u_S(xAy)$ is the median of $\{\mu(y), \mu(x), \sigma(A)\}$ if $\mu(x) \geq \mu(y)$, and the median of $\{\mu(y), \mu(x), \sigma(\overline{A})\}$ otherwise. In order to explain the intuition behind this expression consider the case when someone goes to a meeting by car and has to choose a route. Assume an act $xAy$ comes down to choosing a route such that one gets to the meeting with a huge delay if a traffic jam occurs and with a small delay otherwise. So, $x$ means "arriving to the meeting with little delay", $y$ means "arriving to the meeting with a huge delay", and $A$ means "no traffic jam". Clearly, adopting $u_S(xAy)$ as the utility of the act means the following: if the decision maker is confident enough that there is no traffic jam ($\sigma(A)$ is high enough) then he does as if he trusts the delay will be small ($u_S(xAy) = \mu(x)$); if he has some doubt he might get into a traffic jam then the utility of the act reflects this doubt ($u_S(xAy) = \sigma(A)$); if he totally lacks confidence that the road will be free, then he feels as if there will be a traffic jam ($u_S(xAy) = \mu(y)$). More generally the qualitative monotonic utility can be interpreted as a median, which is satisfactory, since it emphasizes the analogy with expected utility which is an average.

**Proposition 3** [Dubois & Prade, p. 134, 1980] *If $X$ has n+1 elements $\{x_0 = x_*, \ldots, x_n = x^*\}$ with $\mu(x_0) \leq \mu(x_1) \leq \mu(x_{n-1}) \leq \mu(x_n)$, then $u_S(\mathbf{f})$ is the median of the $2n+1$ numbers $\{\sigma(F_x), x \in X, x \neq x_*\} \cup \mu(X)$.*

Then, let us check which of the axioms proposed in the preceding section are satisfied by the monotonic qualitative utility function:

1. **Sav 1** is satisfied, as $u_S$ is an application from $\mathcal{F}$ to a totally ordered scale $L$.
2. Monotonic qualitative utility does not satisfy the sure-thing principle **Sav 2**. It is easy to show that there exist **f, g, h** and **h'** such that $\mathbf{f}A\mathbf{h} \prec \mathbf{g}A\mathbf{h}$, and $\mathbf{f}A\mathbf{h'} \succ \mathbf{g}A\mathbf{h'}$, which is in contradiction with **Sav 2**. Let $B, C, D, E$ be disjoint nonempty subsets of $S$. Let $\sigma$ be a Sugeno measure such that $\sigma(B \cup D) < \sigma(C \cup D)$ and $\sigma(B \cup E) > \sigma(C \cup E)$. Nothing prevents to find such a Sugeno measure, since neither $C \cup D \subseteq B \cup D$ nor $B \cup E \subseteq C \cup E$. The failure of **Sav 2** should not be surprising since Sugeno measures encompasses very different types of set functions, some of them violating a weak form of the additivity property, namely if $(A \cup B) \cap C = \emptyset$, $A >_L B \Rightarrow A \cup B \geq_L B \cup C$. For instance, there are belief functions [Shafer 1976] which violate this property (e.g., with two focal elements $E, F$ such that $m(E) = \alpha < 1 - \alpha = m(F)$; $E \subseteq A$, $F \subseteq B \cup C$ with $F \cap B \neq \emptyset$, $F \cap C \neq \emptyset$ and $A \cap B = \emptyset$).

3. The qualitative utility does not satisfy **Sav 3** nor **Sav 4** proper, due to the failure of the surething principle. Due to the form of $u_S(xAz)$ and $u_S(yAz)$, it is easy to find $x, y, z, A$, such that $u_S(xAz) = u_S(yAz)$ and $x >_P y$, which contradicts **Sav 3**. A similar counterexample may be found that contradicts **Sav 4**.

However our qualitative utility satisfies a property weaker than **Sav 3**:

**WS 3** *Weak compatibility with constant acts:* Let **x** and **y** be constant acts ($\mathbf{x}= x$, $\mathbf{y}= y$), $\forall B \subseteq S$ and $\forall\, \mathbf{h}$, $x \leq_P y \Rightarrow \mathbf{x}B\mathbf{h} \preceq \mathbf{y}B\mathbf{h}$.

To see that the monotonic qualitative utility satisfies **WS 3**, let us introduce a new ordering between acts, that is similar to Pareto dominance in multicriteria decision-making:

**Definition 5** *Pointwise preference:* $\mathbf{f} \leq_P \mathbf{g} \Leftrightarrow \forall s \in S, \mathbf{f}(s) \leq_P \mathbf{g}(s)$.

Pointwise preference is an extension of the total preorder $\leq_P$ on $X$ to a partial preorder $\leq_P$ on $X^S$. Let $X = \{x_0 <_P \ldots <_P x_n\}$ be the set of possible consequences, let **f** be an act, and let $F_n \subseteq \ldots \subseteq F_1 \subseteq F_0 = S$ be the subsets of $S$ such that $F_i = \{s \in S, \mathbf{f}(s) \geq x_i\}$. In the terminology of fuzzy sets, pointwise preference corresponds to fuzzy set inclusion. The monotonicity of the qualitative utility is expressed by the following lemma:

**Lemma 1** $\mathbf{f} \geq_P \mathbf{g} \Rightarrow u_S(\mathbf{f}) \geq u_S(\mathbf{g})$.

This result is already known: it says that Sugeno integral is monotonic with respect to fuzzy set inclusion. Regarding **WS 3**, if $x \leq_P y$, then $\forall A \subseteq S, \forall\, \mathbf{h}$, $\mathbf{x}A\mathbf{h} \leq_P \mathbf{y}A\mathbf{h}$. By Lemma 1, we get $u_S(\mathbf{x}A\mathbf{h}) \leq u_S(\mathbf{y}A\mathbf{h})$, so **WS 3** is satisfied. Our qualitative utility also satisfies a weaker property than **Sav 4**:

**Sav 4'** Let $x >_P x', y >_P y'$ ; $A, B \subseteq S$ : $xAx' \prec xBx' \Rightarrow yAy' \preceq yBy'$. If furthermore $x \geq_P y > y' \geq x'$ then $yAy' \prec yBy' \Rightarrow xAx' \prec xBx'$.

At this point it is important to notice that conditioning on events no longer defines a complete preorder on actions, as $u_S$ does not respect **Sav 2** and **Sav 3**. This is of course due to the fact that $\sigma$ is not necessarily additive.

4. Monotonic qualitative utility satisfies two properties that expected utility does not respect. First, given two acts **f** and **g**, define the act $\mathbf{f} \wedge \mathbf{g}$ (resp. $\mathbf{f} \vee \mathbf{g}$) which in each state $s$ gives the worst (resp. best) of the results $\mathbf{f}(s)$ and $\mathbf{g}(s)$, following the ordering on $X$ (induced by the ordering of constant acts). In terms of fuzzy sets this is the fuzzy union and intersection of fuzzy sets viewed as acts. Due to the pointwise preference lemma, $u_S(\mathbf{f} \wedge \mathbf{g}) \leq min(u_S(\mathbf{f}), u_S(\mathbf{g}))$, and

$u_S(\mathbf{f} \vee \mathbf{g}) \geq max(u_S(\mathbf{f}), u_S(\mathbf{g}))$.

As a consequence of Prop. 3, we prove a weak form of decomposability of the monotonic utility:

**Lemma 2** *Let $\mathbf{f}$ be an act and $\mathbf{y}$ be a constant act of value $y$. Then $u_S(\mathbf{f} \wedge \mathbf{y}) = u_S(\mathbf{f})$ or $\mu(y)$.*

This property indicates that the effect of decreasing the utilities of the best consequences of an act by putting an upper bound to them does not affect the utility of the act until a point where the decision maker starts neglecting the uncertainty pervading this act, and considers that the utility of the act directly reflects this upper bound.

Suppose that you are proposed a lottery, whose prize is a trip to the seaside. As you like the seaside, you are interested in this game. Now, you learn that the prize is changed into a (less preferred) mountain trip ($m$). If $\mathbf{f} = seaW0$ (if you win ($W$), you go to the sea, else you get nothing ($0$)), then the modified game is $\mathbf{g} = \mathbf{f} \wedge m = mW0$. If you are a Sugeno-like DM, either $\mathbf{g} \sim \mathbf{f}$ (you think that winning is not plausible, so, changing the prize does no harm), or $\mathbf{g} \sim m$ (you think that winning is plausible, so that you identify $\mathbf{f}$ and $\mathbf{g}$ with the prizes). Instead, if you are an expected utility maximizer, both $\mathbf{g} \prec \mathbf{f}$ and $\mathbf{g} \prec m$ hold : you do not focus on plausible states, but you make an average between the possible consequences, acting as if the decision problem is faced repeatedly.

It leads to the introduction of a property which is respected by qualitative monotonic utility and not generally by expected utility:

**RCD** *Restricted conjunctive-dominance*: Let $\mathbf{f}$ and $\mathbf{g}$ be any two acts and $\mathbf{y}$ be a constant act of value $y$: $\mathbf{g} \succ \mathbf{f}$ and $\mathbf{y} \succ \mathbf{f} \Rightarrow \mathbf{g} \wedge \mathbf{y} \succ \mathbf{f}$.

**Proposition 4** $u_S$ *satisfies* **RCD**.

**RCD** means that limiting the expectations of an act $\mathbf{g}$ better than another act $\mathbf{f}$ by a constant value that is better than the utility of act $\mathbf{f}$ still yields an act better than $\mathbf{f}$.

To see that expected utility violates **RCD**, it is enough to find real values $a, b, a', b', c$ and a number $\alpha$ in the unit interval such that: $a \cdot \alpha + b \cdot (1 - \alpha) > a' \cdot \alpha + b' \cdot (1-\alpha)$, $c > a' \cdot \alpha + b' \cdot (1-\alpha)$, and $\min(a,c) \cdot \alpha + \min(b,c) \cdot (1-\alpha) < a' \cdot \alpha + b' \cdot (1-\alpha)$. The reader can check that $a = 1000, b = 2, a' = 3, b' = 100, c = 10$, and $\alpha = 0.93$ yields such a counterexample.

**RCD** allows a partial decomposability of the qualitative monotonic utility with respect to the conjunction of acts in the case where one of the acts is constant: $u_S(\mathbf{f} \wedge \mathbf{g}) = \min(u_S(\mathbf{f}), u_S(\mathbf{g}))$ if $\mathbf{g}$ is a constant act, as a consequence of Lemma 2.

There is a dual property that holds for the disjunction of two acts, one of which is a constant act:

**RDD** *Restricted max-dominance*: Let $\mathbf{f}$ and $\mathbf{g}$ be any two acts and $\mathbf{y}$ be a constant act of value $y$: $\mathbf{f} \succ \mathbf{g}$ and $\mathbf{f} \succ \mathbf{y} \Rightarrow \mathbf{f} \succ \mathbf{g} \vee \mathbf{y}$.

**RDD** states that if an act $\mathbf{f}$ is preferred to an act $\mathbf{g}$ and also to the constant act $\mathbf{y}$ then, even if the worst consequences of $\mathbf{g}$ are improved to the value $y$, $\mathbf{f}$ is still preferred to $\mathbf{g}$. Obviously, expected utility does not satisfy **RDD**. In order to prove that $u_S$ satisfies **RDD**, we first prove a lemma, dual of Lemma 2

**Lemma 3** *Let $\mathbf{f}$ be an act and $\mathbf{y}$ be a constant act of value $y$. Then, $u_S(\mathbf{f} \vee \mathbf{y}) = u_S(\mathbf{f})$ or $\mu(y)$.*

By direct application of Lemma 3, the following proposition holds:

**Proposition 5** $u_S$ *satisfies* **RDD**.

A more general decomposability property of the Sugeno utility can also be pointed out: the max-decomposability for *comonotonic acts*. Two acts $\mathbf{f}$ and $\mathbf{g}$ are said to be comonotonic if and only if $\forall s, s' \in S$, $\mathbf{f}(s) >_P \mathbf{f}(s') \Rightarrow \mathbf{g}(s) \geq_P \mathbf{g}(s')$. This property was introduced by [de Campos, Lamata, & Moral 1991]. In words, two acts are comonotonic if none of them gives a strictly better result in state $s'$ than in state $s$, when the other gives a strictly worse result. Note that a similar result is obtained with Choquet integrals which are not additive in general, but become additive for comonotonic functions. A constant act is, of course, comonotonic with any other act. Then the following property holds if $\preceq$ is represented by a Sugeno utility:

**Proposition 6** *If $\mathbf{f}$ and $\mathbf{g}$ are comonotonic, then they verify:* $u_S(\mathbf{f} \vee \mathbf{g}) = \max(u_S(\mathbf{f}), u_S(\mathbf{g}))$, *and* $u_S(\mathbf{f} \wedge \mathbf{g}) = \min(u_S(\mathbf{f}), u_S(\mathbf{g}))$.

## 4 AXIOMATIZATION OF THE MONOTONIC QUALITATIVE UTILITY FUNCTION

The point of this paragraph is to show that if a preference relation $\preceq$ over acts satisfies the axioms **Sav 1**, **WS 3**, **Sav 5**, **RCD** and **RDD** then it can be represented by a monotonic qualitative utility function. The following lemma will be necessary for the proof:

**Lemma 4** *If $\mathbf{f} \geq_P \mathbf{g}$ then $\mathbf{f} \succeq \mathbf{g}$, under* **Sav 1**, **WS 3**, **Sav 5**.

The following theorem can now be proved:

**Theorem 1** *Axiomatization of the monotonic qualitative utility $u_S(\mathbf{f}) = \max_{x \in X} \min(\mu(x), \sigma(F_x))$: Let*

$\preceq$ *be a preference relation over acts satisfying* **Sav 1**, **WS 3**, **Sav 5**, **RCD**, **RDD**. *Then, there exists a finite qualitative scale* $(L, \leq)$, *a utility function* $\mu : X \to L$, *a Sugeno set function* $\sigma : 2^S \to L$, *such that* $\mathbf{f} \preceq \mathbf{f}' \Leftrightarrow u_S(\mathbf{f}) \leq u_S(\mathbf{f}')$.

The proof then goes in four steps:

**Step 1.** The utility scale is the set of acts quotiented by the indifference relation, and $u(\mathbf{f})$ denotes the equivalence class of $\mathbf{f}$. Construct $\mu$ on $X$, by restricting $\preceq$ which is complete (**Sav 1**) to constant acts.

**Step 2.** Suppose now that $\preceq$ also satisfies **WS 3** and **Sav 5**. Extend $u$ to the set of binary acts $x^*Ax_*$. Let $\sigma(A) = u(x^*Ax_*)$, notice that $\sigma(S) = \mu(x^*)$ (1), $\sigma(\emptyset) = \mu(x_*)$, (2), and from Lemma 4, as $A \subseteq B \Rightarrow x^*Ax_* \leq_P x^*Bx_*$, we have $\sigma(A) \leq \sigma(B)$ (3). So, $\sigma$ is a monotonic measure.

**Step 3.** If furthermore $\preceq$ satisfies **RCD**, we extend $u$ to the acts of the form $xAx_*$, and show that $u(xAx_*) = \min(\mu(x), \sigma(A))$.

**Step 4.** First, we prove that for any two binary acts $xAx_*$ and $yBy_*$, where $B \subseteq A$, **RDD** implies $u((xAx_*) \vee (yBx_*)) = \max(u(xAx_*), u(yBx_*))$. Then we prove that $\forall \mathbf{f}, u(\mathbf{f}) = \max_{x \in X} \min(\mu(x), \sigma(F_x))$, noticing that any act $\mathbf{f}$ is the maximum (pointwise) of the $x_i F_{x_i} x_*$. □

**Remark:** Axioms **RDD** and **RCD** can be replaced by an axiom of *max-min decomposability for comonotonic acts*, which is a counterpart of the additive decomposability for comonotonic acts, satisfied by the Choquet integral:

**CoD** If $\mathbf{f}$ and $\mathbf{g}$ are comonotonic, then $\mathbf{f} \vee \mathbf{g} \succ \mathbf{f} \Rightarrow \mathbf{f} \vee \mathbf{g} \sim \mathbf{g}$, and $\mathbf{f} \wedge \mathbf{g} \prec \mathbf{f} \Rightarrow \mathbf{f} \wedge \mathbf{g} \sim \mathbf{g}$.

It is easy to prove that **CoD** implies **RCD** and **RDD**, because a constant act $\mathbf{y}$ is comonotonic with any act. To prove the converse, just follow the proof of Theorem 1. [de Campos, Lamata, & Moral 1991] and [Ralescu & Sugeno 1996] prove representation theorems for Sugeno integrals on the basis of comonotonicity and Lemma 2 taken as an axiom.

## 5    FROM SUGENO INTEGRALS TO QUALITATIVE POSSIBILISTIC UTILITIES

Two qualitative possibilistic criteria for decision making, $QU^*(\mathbf{f}) = \max_{s \in S} \min(\pi(s), \mu(\mathbf{f}(s)))$ and $QU_*(\mathbf{f}) = \min_{s \in S} \max(n(\pi(s)), \mu(\mathbf{f}(s)))$, where $\pi$ is a possibility distribution were proposed and axiomatized in terms of preference over possibilistic lotteries by [Dubois & Prade 1995, Dubois et al. 1998], and by [Dubois, Prade, & Sabbadin 1997] in terms of preference over acts. These two criteria are qualitative counterparts of expected utility. The second one can be seen as a refinement of the Wald criterion, which proposes that the utility of an act be that of its worst possible consequence. Thus, $QU_*$ is "pessimistic" or "cautious", even if the pessimism is moderated by taking relative possibilities of states into account. $QU_*(\mathbf{f})$ is high only if $\mathbf{f}$ gives good consequences in every rather plausible state. On the other side, $QU^*$ is a mild version of the *max* criterion which is "optimistic", or "adventurous" since an act is supposed to be "good", as soon as there exists a plausible state in which it gives a good result.

Possibility and necessity measures are special kinds of Sugeno measures, and it has already been shown that the expressions $QU^*(\mathbf{f})$ and $QU_*(\mathbf{f})$ can be obtained from the expression of $u_S(\mathbf{f})$ by replacing $\sigma$, either by a possibility measure or a necessity measure [Dubois & Prade, p. 134, 1980, Dubois & Prade 1998].

**Proposition 7** *If $\sigma$ is a possibility measure $\Pi$, then*

$$u_S(\mathbf{f}) = QU^*(\mathbf{f}) = \max_{s \in S} \min(\pi(s), \mu(\mathbf{f}(s))).$$

*If $\sigma$ is a necessity measure $N$, then*

$$u_S(\mathbf{f}) = QU_*(\mathbf{f}) = \min_{s \in S} \max(n(\pi(s)), \mu(\mathbf{f}(s)))$$

*where $n$ is the order reversing function of $L$.*

Note that the interpretation of $u_S(xAy) = \max(\mu(y), \min(\sigma(A), \mu(x)))$ $(y <_P x)$, now depends on the meaning of $\sigma$:

i) if $\sigma(A) = \Pi(A)$, then if the decision-maker (DM) has no reason to believe that $A$ will not occur ($\Pi(A)$ is high enough), then he will act as if $A$ occurs ($u_S(xAy) = \mu(x)$). It is only when he strongly thinks that $A$ is impossible ($\Pi(A)$ is low enough) that he evaluates $xAy$ for the worst. This is the optimistic decision-maker.

ii) If $\sigma(A) = N(A)$, it is only when the DM has strong reasons to believe in $A$, that he acts as if it occurs ($u_S(xAy) = \mu(x)$). If he does not, $N(A) = 0$ and he considers that the worst outcomes $y$ will occur. This is the pessimistic decision-maker.

## 6    AXIOMATIZATION OF OPTIMISTIC AND PESSIMISTIC POSSIBILISTIC UTILITIES

Just remember that possibility measures are max-decomposable, that is $\forall A, B, \Pi(A \cup B) = \max(\Pi(A), \Pi(B))$. Therefore, we just strengthen the **RDD** (restricted disjunctive decomposability) axiom, so that for any two acts $x^*Ax_*$ and $x^*Bx_*$ representing bets on events $A$ and $B$, we have $\max(x^*Ax_*, x^*Bx_*) = x^*(A \cup B)x_*$. The strengthened form of the restricted disjunctive dominance is the following:

**DD** *disjunctive dominance:* $\forall$ **f**, **g**, **h**, **f** $\succ$ **g** and **f** $\succ$ **h** $\Rightarrow$ **f** $\succ$ **g** $\vee$ **h**.

Let us point out that **DD** is an axiom expressing that the decision maker focuses on the "best" plausible states. It is easier to see, if instead of **DD** we take the following equivalent axiom:

**Optimism** $\forall$ **f**, **g**, $\forall A \subseteq S$, **f**A**g** $\prec$ **f** $\Rightarrow$ **f** $\preceq$ **g**A**f**.

**Proposition 8** *DD and Optimism are equivalent.*

Let us see what it means for an agent to obey **Optimism** on the following very simple example: Suppose that **f** is the act that makes you earn a gain $G$ whatever the state of the world, and **g** makes you loose a loss $L$, whatever the state of the world. If you obey **Optimism**, then for any given $A$, **f** is strictly preferred to **f**A**g** means that you think that $\bar{A}$ is rather plausible (if not, you would be indifferent). You do not prefer **f** to **g**A**f** means that you think that either $A$ is not plausible, or it is and your preferences given $A$ are blurred by the fact that the consequences on $\bar{A}$ which is also plausible are very good. In other terms, for making up your mind, you focus on the best consequences among those of the plausible states. This is optimism.

**Proposition 9** $QU^*$ *satisfies* **DD**.

It is sufficient to prove that if $\preceq$ satisfies **Sav 1**, **WS 3**, **Sav 5**, **RCD** and **DD**, then the preorder induced on events is a comparative possibility, in order to prove that $\sigma$ is a possibility measure, and that $\preceq$ can be represented by a qualitative possibilistic optimistic utility function $QU^*$.

**Proposition 10** *Let $\preceq$ be a preorder over $X^S$ satisfying* **Sav 1**, **WS 3**, **Sav 5**, **RCD** *and* **DD**, *and let $\leq_L$ be the corresponding induced order over events:* $\forall A, B \subseteq S, A \leq_L B \Leftrightarrow 1A0 \preceq 1B0$.

$\leq_L$ *is a comparative possibility relation.*

It is easy to show that if $\preceq$ satisfies **Sav 1**, **WS 3**, **Sav 5**, then $\leq_L$ satisfies **A1**, **A2** and **A3**. Then, we have to prove that $\leq_L$ satisfies the characteristic axiom of comparative possibility relations: $\Pi$: $\forall A, B, C, B \leq_L C \Rightarrow A \cup B \leq_L A \cup C$ in order to prove that it is a comparative possibility relation. The proof is trivial, noticing that if $\preceq$ satisfies the axioms, then **f**$\vee$**g** $\sim$ $\max_{\preceq}$(**f**, **g**). Just use bets on events $A, B, C$ to prove the result.

Now, the following theorem is easy to prove:

**Theorem 2** *Let $\preceq$ be a preorder over $X^S$ satisfying* **Sav 1**, **WS 3**, **Sav 5**, **RCD** *and* **DD**. *There exists $L$ a finite totally ordered scale, $\mu : X \rightarrow L$ a utility function and $\pi : S \rightarrow L$ a possibility distribution, such that $\preceq$ is represented by $QU^*$:* **f** $\rightarrow QU^*$(**f**) $= \max_{s \in S} \min(\pi(s), \mu(\mathbf{f}(s)))$.

The axiomatic justification of the pessimistic utility $QU_*$ can be obtained in a dual way: first, we recover the pessimistic qualitative utility by reinforcing **RCD** instead of **RDD**.

**CD** *Conjunctive dominance:* $\forall$ **f**, **g**, **h**, **g** $\succ$ **f** and **h** $\succ$ **f** $\Rightarrow$ **g** $\wedge$ **h** $\succ$ **f**.

In the same way as we did for the qualitative optimistic utility, we can prove the following theorem:

**Theorem 3** *Let $\preceq$ be a preorder over $X^S$ satisfying* **Sav 1**, **WS 3**, **Sav 5**, **RDD** *and* **CD**. *There exists $L$ a finite totally ordered scale, $\mu : X \rightarrow L$ a utility function and $\pi : S \rightarrow L$ a possibility distribution, such that $\preceq$ is represented by $QU_*$:* **f** $\rightarrow QU_*(\mathbf{f}) = \min_{s \in S} \max(n(\pi(s)), \mu(\mathbf{f}(s)))$, $n$ *being the order-reversing function of* $L$

Notice that **CD** can also be represented by a "pessimism" or "cautiousness" axiom dual to the **Optimism** axiom:

**Pessimism**: $\forall$ **f**, **g**, $\forall A \subseteq S$, **f**A**g** $\succ$ **f** $\Rightarrow$ **f** $\succeq$ **g**A**f**.

The same example, obeying **Pessimism** means that if you strictly prefer **f**A**g** to **g** (you think that $A$ is plausible), then you should not prefer **g**A**f** to **g** (the bad consequence $L$ blurs your preferences given $\bar{A}$).

## 7 CONCLUDING REMARKS

The qualitative decision theory outlined here significantly differs from the usual one, based on expected utility. It only presupposes a finite setting and a linear qualitative scale, while expected utility resorts to numerical uncertainty and utility functions, and infinite state space or set of consequences. Moreover it is non-compensatory since it rejects the notion of averaging between the value of uncertain outcomes. Our approach is also very general since the set functions representing the decision-maker's uncertainty need only be monotone. The Sugeno integral representing utility is the qualitative counterpart of Choquet integral, as axiomatized in [Schmeidler 1989, Sarin & Wakker 1992]. Medians substitute to mean values or bounds thereof. If the attitude of the DM in front of uncertainty is taken into account we recover possibilistic utilities, thus getting a milder, more realistic version of Wald criterion. While our results suggest natural decision criteria in the presence of qualitative information and highlight the underlying as-

128    Dubois, Prade, and Sabbadin

sumptions, the axiomatic setting based on acts also gives a tool for observing if a decision-maker's way of representing uncertainty follows the rules of possibility theory. Interestingly similar results have been independently recently obtained in the setting of multicriteria aggregation by [Marichal 1997]. One strong assumption has been made in this paper (and also in classical utility theory), which is that certainty levels and priority levels be commensurate. An attempt to relax this assumption has been made in [Dubois, Fargier, & Prade 1997]. These authors point out that working without the commensurability assumption leads them to the theory of uncertainty underlying preferential entailment in non-monotonic reasoning. Unfortunately, the corresponding decision methods also prove to be either indecisive or very risky.

This paper is only a first step in the direction of a full-fledged qualitative decision theory in the Savage style. The next step should be a proper handling of conditional acts, which is nontrivial in the absence of the Sure thing principle. Some ideas about how to address this question can be borrowed from [Lehmann 1996].